\documentclass{article}

\usepackage[preprint]{neurips_2026}

\usepackage[utf8]{inputenc}
\usepackage[T1]{fontenc}
\usepackage{hyperref}
\usepackage{url}
\usepackage{booktabs}
\usepackage{amsfonts}
\usepackage{amssymb}
\usepackage{amsmath}
\usepackage{nicefrac}
\usepackage{microtype}
\usepackage{xcolor}
\usepackage{graphicx}

\title{Beyond Exponential Decay:\\Rethinking Error Accumulation in Large Language Models}

\author{%
  Mikhail L. Arbuzov \\
  Independent Researcher \\
  \texttt{mike.arbuzov54@gmail.com} \\
  \And
  Sisong Bei \\
  Independent Researcher \\
  \texttt{qurining@gmail.com} \\
  \And
  Ziwei Dong \\
  Independent Researcher \\
  \texttt{ziwei.dong@alumni.emory.edu} \\
  \And
  Dmitri Kalaev \\
  Independent Researcher \\
  \texttt{kalaevdr@gmail.com} \\
  \And
  Alexey Shvets \\
  Palo Alto Networks \\
  \texttt{ashvets@paloaltonetworks.com} \\
}

\begin{document}

\maketitle

\begin{abstract}
A common pessimistic argument holds that autoregressive language models suffer exponential decay in correctness over long outputs: if each token has independent error probability $e$, then $(1-e)^n \to 0$ as $n$ grows. The argument is clean to state and widely cited. It is also brittle, and three lines of recent empirical work make the cracks visible. The first is that only a small subset of tokens---roughly $5\%$ to $10\%$ in the studies that have actually measured it---genuinely depends on long-range context; the rest get more predictable, not less, as context accumulates. The second is geometric: LLM embeddings organize into stratified low-dimensional manifolds, so once a model is working inside one semantic region it tends to stay there even when individual tokens slip. The third concerns what happens when models do err on the consequential tokens---errors turn out to be idiosyncratic across samples rather than systematic, which is why majority-vote ensembles recover so much accuracy. Pulling these together gives a two-rate model, $P(\text{correct}) \approx (1-e_{\text{key}})^k \cdot (1-e_{\text{non}})^{n-k}$, in which $k$ scales sublinearly with $n$ and $e_{\text{non}}$ approaches zero with sufficient context. The predicted decay is, at worst, stretched-exponential; often power-law; and when $k$ saturates at some task-specific $k_{\max}$, constant in $n$. A number of recent capabilities---anchor compression at $99\%$ context reduction, $128$K-token retrieval on consumer GPUs, self-consistency gains on reasoning benchmarks---then read as natural consequences of one structural fact rather than independent engineering wins: long-context reliability hinges on a handful of decision points, not on uniform per-token accuracy.
\end{abstract}

\section{Introduction}
\label{sec:intro}

Autoregressive language models, the argument goes, are doomed for long outputs. If each token has even a $1\%$ error rate then a $100$-token chain is correct only $(0.99)^{100} \approx 37\%$ of the time, and longer chains decay exponentially toward zero \citep{lecun2023why,dziri2023faith}. The argument is clean, and it has been used to predict a hard ceiling on what autoregressive systems can do.

What modern LLMs actually do is harder to square with this picture. Multi-page outputs hold together. Mid-generation self-correction is routine---later tokens revise earlier interpretations, a behavior \citet{gwern2023inner} called out as incompatible with monotonically increasing error. The attention patterns are lopsided: $96\%$ of cumulative attention weight in Llama-3 concentrates on $\sim 1\%$ of the context \citep{liu2024retrieval}, and only about $9\%$ of tokens in natural text show meaningful dependence on distant context \citep{fang2024what}. With the right test-time strategy, a $1$B-parameter model can match or beat a $405$B model on specific reasoning tasks \citep{venture2025test}. None of this is consistent with uniform per-token error.

Our reading is that the independent-error model fails because it averages over a heterogeneous population of tokens. A small fraction---key tokens---carry the load of long-range dependency and global coherence. The rest are constrained by local syntax and the accumulating context, and their error rate goes to zero rather than to a constant. Three threads of recent work make this concrete: \citet{fang2024what} on which tokens actually need long context, \citet{li2025unraveling} on the stratified-manifold organization of embeddings, and \citet{wang2023self} on the convergence of correct reasoning paths under sampling. Combined, they produce a refined model:
\begin{equation}
\label{eq:two-rate}
P(\text{correct}) \approx (1-e_{\text{key}})^k \cdot (1-e_{\text{non}})^{n-k},
\end{equation}
where $k$ is the number of key tokens, $e_{\text{key}}$ their error rate, and $e_{\text{non}}$ the much lower error rate for the remaining $n-k$. Because $k$ scales sublinearly with $n$---logarithmically, as a fractional power, or saturating at some $k_{\max}$---and because $e_{\text{non}} \to 0$ with context, the predicted decay is far gentler than $(1-e)^n$.

\paragraph{Contributions.}
(1)~A two-rate error model that distinguishes key from non-key tokens and makes the dependence on $k(n)$ explicit (\S\ref{sec:framework}). (2)~A synthesis of evidence from three recent research streams---attention sparsity, embedding geometry, and ensemble convergence---that supports each component of the model (\S\ref{sec:evidence}). (3)~An accounting of how this framework reframes existing systems-level results (anchor compression, retrieval-augmented attention, self-consistency, tool integration) as predictable consequences of a single structural fact rather than independent engineering wins (\S\ref{sec:implications}).

\section{Related Work}
\label{sec:related}

\paragraph{Error compounding.}
The exposure-bias problem is old. \citet{bengio2015scheduled} formalized it: models trained with teacher forcing must, at inference time, condition on their own potentially noisy output, so small errors can compound through state updates. \citet{lecun2023why} extended the worry to autoregressive LLMs in the now-familiar form---if each token has error probability $e$, sequence-level correctness decays as $(1-e)^n$. The bound is tight when independence really holds, and loose otherwise; transformers attending backward over their own output is exactly the case where it goes loose, and Gwern's inner-monologue evidence \citep{gwern2023inner} of error rates that \emph{decrease} at certain points in a sequence is hard to reconcile with strict monotonic accumulation. We keep the per-token decomposition but partition tokens by their long-range dependency.

\paragraph{Long-context utilization.}
\citet{liu2023lost} documented the lost-in-the-middle effect---models disproportionately attend to material near the beginning and end of the prompt. The same skew was given a sharper edge by \citet{fang2024what}, whose long-short difference (LSD) metric finds only $\sim 9\%$ of tokens in natural text scoring LSD$>2$. \citet{liu2024retrieval} showed $96\%$ of cumulative attention weight concentrating on $\sim 1{,}000$ tokens out of $100{,}000$ in Llama-3, and used the observation to enable $128$K-token contexts on consumer GPUs. Anchor compression \citep{pang2024anchor} reaches $99\%$ context reduction with $<1.5\%$ accuracy loss. Each of these papers treats sparsity as an empirical fact about a particular system; we read them as direct evidence for $k \ll n$.

\paragraph{Embedding geometry.}
\citet{li2025unraveling} probed LLM embedding spaces with sparse mixture-of-experts and found a stratified-manifold structure: a union of low-dimensional submanifolds aligned with semantic domain. The story gets more concrete with \citet{robinson2025token}, who distinguished signal from noise dimensions in token embeddings---perturbing signal dimensions reroutes outputs, perturbing noise dimensions does not. \citet{viswanathan2024geometry} link intrinsic dimensionality to prediction loss; high-confidence contexts collapse to lower-dimensional representations. And \citet{gao2023transformers}, probing intermediate layers, recover the correct answer from the model's internals more than $80\%$ of the time even when the surface output was wrong. Compartmentalized representations, signal/noise separation, dimensional collapse under confidence: this is what lets the model absorb the small token-level slips without breaking coherence at the semantic level.

\paragraph{Ensemble reasoning.}
Self-consistency, in \citet{wang2023self}'s formulation, is almost embarrassingly simple: sample multiple reasoning paths, take the majority answer. The gain was $+17.9$ points on GSM8K, with no retraining. Tree-of-Thoughts \citep{yao2023tree} extends the idea with structured tree search at high-uncertainty branch points and pulls GPT-4 on Game-of-24 from $4\%$ to $74\%$. \citet{li2023reliability} sort the effects by error type, separating systematic errors (knowledge gaps---similar across samples) from idiosyncratic ones (variable across samples), and that distinction explains why ensembles work for reasoning but not for retrieval. The size of these effects is the part the independent-error model cannot absorb. If errors compounded uniformly, multiple samples would yield multiple failures, not a more accurate mode.

\section{Theoretical Framework}
\label{sec:framework}

We develop the two-rate model in three stages: the partition of tokens into key and non-key, the manifold dynamics that structure error correlation, and the redundancy gain that ensemble methods extract.

\subsection{Two-Rate Error Model}

The independent-error argument assumes every token carries equal weight in the failure probability. It does not. Key tokens are the ones whose correctness genuinely depends on long-range context or global knowledge---factual claims, logical operators, points of co-reference, transitions between topics. Non-key tokens are governed by local regularity: syntax, frequent collocations, content already established by the surrounding text. Their error rate is small and decreases as more context accumulates.

In a sequence of length $n$, let $k$ count the key tokens. The empirical estimate of $k/n$ from \citet{fang2024what} is $\sim 9\%$; adversarial-perturbation studies converge on a similar range \citep{morris2022textattack}. We hypothesize that $k$ grows sublinearly with $n$, possibly saturating at a task-specific $k_{\max}$: even a book-length argument operates within a finite knowledge frame requiring a bounded number of critical decisions. With per-token error rates $e_{\text{key}}$ (large) and $e_{\text{non}}$ (small, decreasing in context), Eq.~\ref{eq:two-rate} gives three regimes:

\begin{enumerate}
\setlength{\itemsep}{1pt}
\item \textbf{Logarithmic key-token growth} ($k \sim \log n$): polynomial decay $n^{-c}$, much slower than exponential.
\item \textbf{Fractional-power growth} ($k \sim \sqrt{n}$): stretched-exponential decay.
\item \textbf{Saturating} ($k \to k_{\max}$): reliability becomes constant in $n$ once the key facts are committed.
\end{enumerate}

The third regime is the strange one. \citet{pang2024anchor} achieved $99\%$ context reduction with $<1.5\%$ accuracy loss---a result that requires $k$ to be effectively bounded for the relevant tasks. There is no way to express that under exponential decay; the functional form forbids it.

\subsection{Stratified Manifold and Error Correlation}

The $(1-e)^n$ form also assumes errors are independent across positions. They are not. The embedding space's stratified structure \citep{li2025unraveling} introduces correlation, and we can read the dynamics off it directly. Treat the model's hidden state as moving along a manifold $M_C$ determined by the prevailing context. Most errors are minor---a synonym substitution, a grammatical slip, a momentary wobble that stays on $M_C$ and has no downstream effect. Disruptive errors are different: a key-token mistake jumps the trajectory onto a different manifold $M_{C'}$, after which subsequent tokens cohere with the wrong commitment. The result is a fluent-but-wrong continuation. Errors cluster rather than scatter.

This correlation structure cuts the union-bound failure probability. For small $k$,
\begin{equation}
P(\text{any disruptive error}) \leq k \cdot e_{\text{key}},
\end{equation}
which, when $k \ll n$, is substantially below $1-(1-e)^n$. The manifold structure also explains the compartmentalization \citet{gao2023transformers} document: intermediate layers preserve correct internal representations even when the surface output strays, because the hidden-state trajectory remains on the correct manifold even when token sampling does not.

\subsection{Self-Consistency and Redundancy Gain}

When key-token errors are idiosyncratic across samples---different paths fail at different junctions---majority-vote ensembles can recover the correct answer. For $m$ samples with error correlation $\rho$,
\begin{equation}
e_{\text{key}}^{\text{(eff)}} = f(\rho, m) \cdot e_{\text{key}},
\end{equation}
where $f \to 1$ at $\rho = 1$ (perfect correlation, no gain) and $f \to e_{\text{key}}^{m-1}$ at $\rho = 0$ (independent errors, exponential gain). Wang et al.'s self-consistency results \citep{wang2023self} sit closer to the $\rho = 0$ regime than to $\rho = 1$, which is why a method that adds no parameters and no training data can lift GSM8K accuracy by $17.9$ points: the gain is structural, not a property of the sampling temperature.

\citet{costello2025think} push this further: as a model's Pass@1 accuracy improves through iterative self-training, the marginal gain from majority voting falls. Correct reasoning paths cluster on a narrow manifold; incorrect ones fan out. Ensemble methods only help when the model's spread of guesses straddles the right region.

\section{Empirical Evidence}
\label{sec:evidence}

Three predictions, three lines of evidence. The framework says $k/n$ is small and stable; that token errors correlate via manifold geometry; and that correct reasoning paths converge while incorrect ones diverge. The supporting work for each comes from a different methodology, which is what makes the convergence interesting.

\subsection{Key-Token Sparsity}

The most direct measurement is \citet{fang2024what}'s LSD metric: $9\%$ of tokens in natural text qualify as key (LSD$>2$), and perplexity restricted to those tokens correlates with downstream task performance at Pearson $\approx -0.96$. Perplexity on the other $91\%$ tracks task success at essentially zero correlation. Adversarial robustness work, working from the opposite direction, lands in the same band---\citet{morris2022textattack} flip model decisions by perturbing $5\%$--$10\%$ of strategically chosen tokens, while random perturbation of much larger fractions does nothing. The $5\%$--$10\%$ figure recurs across methodologies built to measure different things; that is what one would expect if it were a structural property of natural language rather than a dataset artifact.

The systems-level corollary is sharp. \citet{liu2024retrieval} reach $128$K-token effective context on consumer GPUs by computing attention only over the top $\sim 1{,}000$ tokens by attention mass, recovering $>90\%$ of full-attention scores in the process. Anchor-LLM \citep{pang2024anchor} compresses sequence information into a single token at $99\%$ reduction with $<1.5\%$ accuracy loss. Neither method would work if token importance were uniform; both work because it is not.

\subsection{Stratified Manifold Structure}

\citet{li2025unraveling}'s sparse-MoE probe of frozen LLM embeddings finds a union of low-dimensional manifolds aligned with semantic domain (scientific text, dialogue, code), with measurably different intrinsic dimension across regions. \citet{robinson2025token}'s fiber-bundle analysis distinguishes regular-neighborhood tokens (manifold interior; perturbation has minimal effect) from irregular-neighborhood tokens (manifold junctions; perturbation reroutes generation). The mapping to our key/non-key partition is direct: irregular-neighborhood tokens are exactly the points where a small input change can move the trajectory between manifolds.

\citet{viswanathan2024geometry} link this to confidence: prompts that elicit low-intrinsic-dimensional representations correlate with lower prediction loss. The model's state contracts as confidence rises---which is why the per-token error rate $e_{\text{non}}$ should fall, not stay constant, as context accumulates.

\citet{gao2023transformers} probe intermediate layers and recover correct answers from $>80\%$ of cases where the final output was wrong. The information was preserved internally; only the surface form failed. This is the manifold-stays-correct behavior our framework predicts.

\subsection{Convergent Reasoning Paths}

\citet{wang2023self} showed that self-consistency lifts GSM8K by $17.9$ points and SVAMP by $11.0$, with no model change. The mechanism, on our reading, is that correct paths concentrate near a low-dimensional attractor while incorrect ones disperse---so majority voting picks out the attractor's mode. \citet{yao2023tree}'s Tree-of-Thoughts moves GPT-4 on Game-of-24 from $4\%$ to $74\%$ by branching at high-uncertainty points specifically. \citet{li2023reliability} explain why these methods work for reasoning and not for knowledge retrieval: knowledge gaps produce systematic errors (every sample fails the same way); reasoning slips produce idiosyncratic ones (samples fail differently). The framework's two error rates accommodate both: $e_{\text{key}}$ rises uniformly when the model lacks the relevant fact, and ensembling cannot help.

A summary of how the systems-level results map onto the framework's three pillars appears in Appendix~\ref{app:systems}.

\section{Practical Implications}
\label{sec:implications}

The framework is not just a description of why $(1-e)^n$ is wrong. It also predicts where computation should go, how context should be managed, and what to evaluate.

\paragraph{Sparse attention and context compression.}
If $k \ll n$, attention does not need to be dense. \citet{pang2024anchor}'s anchor compression and \citet{liu2024retrieval}'s top-$k$ attention are the existing demonstrations; \citet{wu2024tokenselect}'s dynamic KV-cache selection is another point on the same curve. These methods stop being clever tricks and become predictable: dense attention pays a quadratic cost to recover a sparse signal.

\paragraph{Targeted compute at decision points.}
Errors concentrate at manifold transitions, so compute should too. \citet{moshkov2025aimo}'s tool-integrated reasoning fires Python execution at high-entropy spans rather than uniformly. Adaptive computation time \citep{xin2023adaptive} lets confident tokens skip layers altogether, saving $40\%$--$60\%$ on SST-2 and TriviaQA. The adaptive-temperature decoding of \citet{zhu2024hotorcold} raises exploration at uncertain tokens and damps it elsewhere. Three different prescriptions, one underlying instruction: spend cycles where the manifold is about to fork.

\paragraph{Strategic ensembles.}
\citet{wang2023self}'s self-consistency, \citet{yao2023tree}'s tree search, and \citet{moshkov2025aimo}'s GenSelect all exploit the convergence-of-correct-paths structure. The framework explains why they work---and predicts when they will not, namely when errors are systematic rather than idiosyncratic, as in pure knowledge-retrieval tasks \citep{li2023reliability}.

\paragraph{Evaluation aligned with key tokens.}
Plain perplexity averages over a population the model is trying to handle separately, which is why it underperforms as a predictor. \citet{fang2024what}'s LongPPL restricts perplexity to key tokens and lifts the correlation with downstream performance to $r = -0.96$ from a near-zero baseline. \citet{costello2025think}'s success-plateau curves show what reliability looks like up close: sharp drops after extended plateaus---the staircase one would expect under our model, not the smooth exponential decay of the alternative.

A more speculative implication---modular reasoning architectures that route by manifold region, building on the alignment-not-scale results of \citet{costello2025think}---we defer to Appendix~\ref{app:arch}.

\section{Limitations}
\label{sec:limitations}

The framework is a synthesis, not a derivation. Three cautions belong on record. The two-rate model is descriptive: $k$, $e_{\text{key}}$, and $e_{\text{non}}$ are observable in principle but not yet jointly measured on a single benchmark, so the predicted decay regimes are arguments from supporting evidence rather than fitted curves. The manifold-structure account leans on \citet{li2025unraveling} and \citet{robinson2025token} for direct geometric evidence; both are recent, and replication on larger model scales would tighten the claim. The convergent-paths claim holds where \citet{li2023reliability} call errors idiosyncratic; we have no quantitative criterion for distinguishing the idiosyncratic from the systematic regime in advance, only the post-hoc observation that ensemble methods help in one and not the other.

\section{Conclusion}
\label{sec:conclusion}

The independent-error argument was always going to fail in one of two places. Either some tokens would matter more than others, breaking the per-token uniformity; or errors would correlate, breaking the independence. As it turns out, both fail at once. The empirical literature has been steadily accumulating the evidence: $5\%$--$10\%$ of tokens carry the long-range dependency, embeddings stratify into manifolds that compartmentalize errors, and ensemble methods extract gains that any uniform-error model rules out.

The two-rate model in Eq.~\ref{eq:two-rate} is a small change to the algebra and a large change to the predicted behavior. Reliability depends on $k$ key decisions, not on $n$ tokens; and $k$ scales sublinearly, often saturating. Modern LLMs hold coherence across thousands of tokens not because the underlying $(1-e)^n$ is being beaten by clever engineering, but because $(1-e)^n$ was the wrong functional form to begin with.

The systems-level consequences are already visible. Anchor compression, retrieval-augmented attention, tool-integrated reasoning, and self-consistency all gain explanatory unity once they are read as instances of a single principle: identify where the manifold forks, and put the compute there. Future work should make $k(n)$ measurable on a fixed benchmark, derive tighter bounds from attention-pattern statistics rather than from heuristic union bounds, and connect token-level decay to the reasoning-step decay that bears more directly on agentic systems.

\bibliographystyle{plainnat}
\bibliography{references}

\appendix

\section{Architectural Implications: Modular Reasoning}
\label{app:arch}

The stratified-manifold view suggests an architectural prescription the main body only gestures at: instead of scaling monolithic models, route reasoning subtasks to specialized models aligned with manifold regions. The clearest existing evidence comes from alignment-not-scale work. \citet{costello2025think}'s Trace-Prune-Train pipeline shows that smaller models ($2$--$9$B parameters) iteratively fine-tuned on their own pruned reasoning traces can match models $30\times$ larger on GSM8K---improving Gemma2-2B from $41.9\%$ to $57.6\%$ Pass@1, Gemma2-9B to $82\%$ (matching LLaMA-3.1-70B), and LLaMA-3.1-70B to $91\%$ (above GPT-4o's $82\%$). Whether explicit routing on top of these aligned smaller models compounds the gains, or runs into a ceiling once the routed-to manifold is itself stratified, is the natural next question; we are not aware of a definitive empirical answer.

The pattern across these systems is consistent. Fitting the manifold beats expanding the parameter count, when the task population is narrow enough that the manifold is well-defined. The open question is how broad a domain a single specialized model can cover before its internal stratification reasserts itself and the routing problem reappears one level down. We do not attempt to settle that here; it is a question for empirical work that systematically varies routing granularity.

\section{Extended Case Studies in Advanced Reasoning Systems}
\label{app:systems}

The main body summarizes the systems-level evidence at a high level. Three case studies bear closer reading.

\paragraph{AIMO-2 (\citet{moshkov2025aimo}).}
The winning solution to the AI Mathematical Olympiad combines all three pillars of the framework. Tool integration fires at high-entropy spans, addressing key-token uncertainty directly; the dataset-creation pipeline exploits the structured nature of correct reasoning paths to generate training data; GenSelect leverages the convergent-paths property to pick the best candidate solution from many. The system reaches state-of-the-art on AIME and Harvard-MIT Mathematics Tournament problems. Most striking, the GenSelect mechanism works on \emph{compressed summaries} of reasoning traces, which would be impossible if errors were spread uniformly across the full output: the summaries would lack the discriminating signal.

\paragraph{Test-time compute scaling (\citet{nvidia2023scaling,venture2025test}).}
Optimal test-time strategy depends on problem difficulty and model size. Beam search dominates best-of-$N$ on hard problems for sub-$7$B models; the order reverses on easier problems for larger models. The framework reads this as a direct consequence of $k$ varying with task: harder tasks have more decision points to traverse correctly, rewarding search-based exploration; easier ones have few, rewarding sampling-based ensembling. The two-orders-of-magnitude result---an optimized $1$B model outperforming a $405$B model on specific reasoning tasks \citep{venture2025test}---is exactly what an independent-error model rules out and what the two-rate model predicts under saturation.

\paragraph{Attention concentration in production models (\citet{liu2024retrieval}).}
Llama-3 places $96\%$ of cumulative attention weight on $\sim 1{,}000$ tokens out of $100{,}000$. RetrievalAttention exploits this directly: at generation time, only the top-mass tokens enter attention, with the rest treated as a nearest-neighbor lookup. The result is $128$K effective context on a single RTX 4090. The structural claim is the same one $k \ll n$ encodes; the engineering is just paying attention to it.

A consolidated summary of how each result maps onto the framework's three pillars appears in Table~\ref{tab:implications}.

\begin{table}[ht]
\caption{How systems-level results map onto the framework's three pillars.}
\label{tab:implications}
\centering
\small
\begin{tabular}{p{2.6cm}p{3cm}p{3cm}p{3.5cm}}
\toprule
\textbf{Challenge} & \textbf{Independent-error view} & \textbf{Two-rate view} & \textbf{Exemplar systems} \\
\midrule
Long-context handling & Uniform attention over all tokens & Sparse retrieval focused on key tokens & Anchor LLMs \citep{pang2024anchor}; RetrievalAttention \citep{liu2024retrieval} \\
\midrule
Compute allocation & Equal resources for all tokens & Targeted at manifold transitions & Tool integration \citep{moshkov2025aimo}; ACT \citep{xin2023adaptive} \\
\midrule
Error reduction & Independent samples, multiplicative gain & Branching at uncertain junctions & Self-consistency \citep{wang2023self}; GenSelect \citep{moshkov2025aimo} \\
\midrule
Evaluation & Uniform perplexity & Key-token-restricted metrics & LongPPL \citep{fang2024what}; success-plateau curves \citep{costello2025think} \\
\midrule
Architecture & Scale monolithic models & Alignment over scale & TPT \citep{costello2025think} \\
\bottomrule
\end{tabular}
\end{table}
\end{document}